\documentclass{article}

\usepackage{amsmath,amsfonts,bm}

\def\eqref#1{equation~\ref{#1}}

\def\1{\bm{1}}

\DeclareMathAlphabet{\mathsfit}{\encodingdefault}{\sfdefault}{m}{sl}
\SetMathAlphabet{\mathsfit}{bold}{\encodingdefault}{\sfdefault}{bx}{n}

\usepackage[legalpaper, margin=1in]{geometry}
\usepackage{authblk}
\usepackage{hyperref}
\usepackage{url}
\usepackage{graphicx}
\usepackage{subcaption}
\usepackage{algorithm}
\usepackage{algorithmic}
\usepackage{natbib}
\usepackage{color}
\usepackage{listings}
\usepackage{xcolor}

\definecolor{codegreen}{rgb}{0,0.6,0}
\definecolor{codegray}{rgb}{0.5,0.5,0.5}
\definecolor{codepurple}{rgb}{0.58,0,0.82}
\definecolor{backcolour}{rgb}{0.95,0.95,0.92}
\definecolor{weborange}{RGB}{255,165,0}
\definecolor{frenchplum}{RGB}{129,20,83}
\definecolor{carrotorange}{rgb}{0.93, 0.57, 0.13}

\lstdefinestyle{mystyle}{
    language=Python,
    commentstyle=\color{codegreen},
    keywordstyle=\color{blue},
    numberstyle=\tiny\color{codegray},
    stringstyle=\color{codepurple},
    basicstyle=\ttfamily\footnotesize,
    emph={kernel, stride, padding},
    emphstyle={\color{carrotorange}},
    breakatwhitespace=false,         
    breaklines=true,                 
    captionpos=b,                    
    keepspaces=true,                 
    numbers=left,                    
    numbersep=5pt,                  
    showspaces=false,
    showstringspaces=false,
    showtabs=false,                  
    tabsize=2,
    otherkeywords={Sequential, Conv2d, AvgPool2d, LayerNorm, Linear, Flatten},
    morekeywords={Sequential, Conv2d, AvgPool2d, LayerNorm, Linear, Flatten},
}

\title{FedEmbed: Personalized Private Federated Learning}

\author[1]{Andrew Silva\footnote{Work done as an intern at Apple.}}
\author[2]{Katherine Metcalf}
\author[2]{Nicholas Apostoloff}
\author[2]{Barry-John Theobald}
\affil[1]{Georgia Institute of Technology}
\affil[2]{Apple}
\date{}     

\renewenvironment{abstract}
{\small
\begin{center}
\bfseries \abstractname\vspace{-.5em}\vspace{0pt}
\end{center}
\list{}{
\setlength{\leftmargin}{2cm}%
\setlength{\rightmargin}{\leftmargin}%
}%
\item\relax}
{\endlist}

\begin{document}

\maketitle

\begin{abstract}
Federated learning enables the deployment of machine learning to problems for which centralized data collection is impractical. Adding differential privacy guarantees bounds on privacy while data are contributed to a global model. Adding personalization to federated learning introduces new challenges as we must account for preferences of individual users, where a data sample could have conflicting labels because one sub-population of users might view an input positively, but other sub-populations view the same input negatively. We present FedEmbed, a new approach to private federated learning for personalizing a global model that uses (1) sub-populations of similar users, and (2) personal embeddings. We demonstrate that current approaches to federated learning are inadequate for handling data with conflicting labels, and we show that FedEmbed achieves up to 45\% improvement over baseline approaches to personalized private federated learning. \vspace{1cm}
\end{abstract}

\section{Introduction}
\label{sec:intro}
 As the scale and diversity of problems tackled with machine learning increases, so too does the challenge of acquiring sufficiently large, centralized, real-world datasets. For example, legal restrictions may explicitly restrict or prohibit collecting centralized datasets containing personally identifiable information \citep{roberts2017evaluating,gerke2019ethical,gerke2020ethical}. Federated learning \citep{mcmahan2017communication} learns from \textit{decentralized} data to address many of the challenges associated with centralized datasets: clients obtain a model from a server and perform gradient updates on their local data; the gradients are sent to the server; the server updates the model, e.g., using \textit{federated averaging}; then the updated model is pushed to all clients.

\begin{figure}[t]
    \centering
    \includegraphics[width=1.0\linewidth]{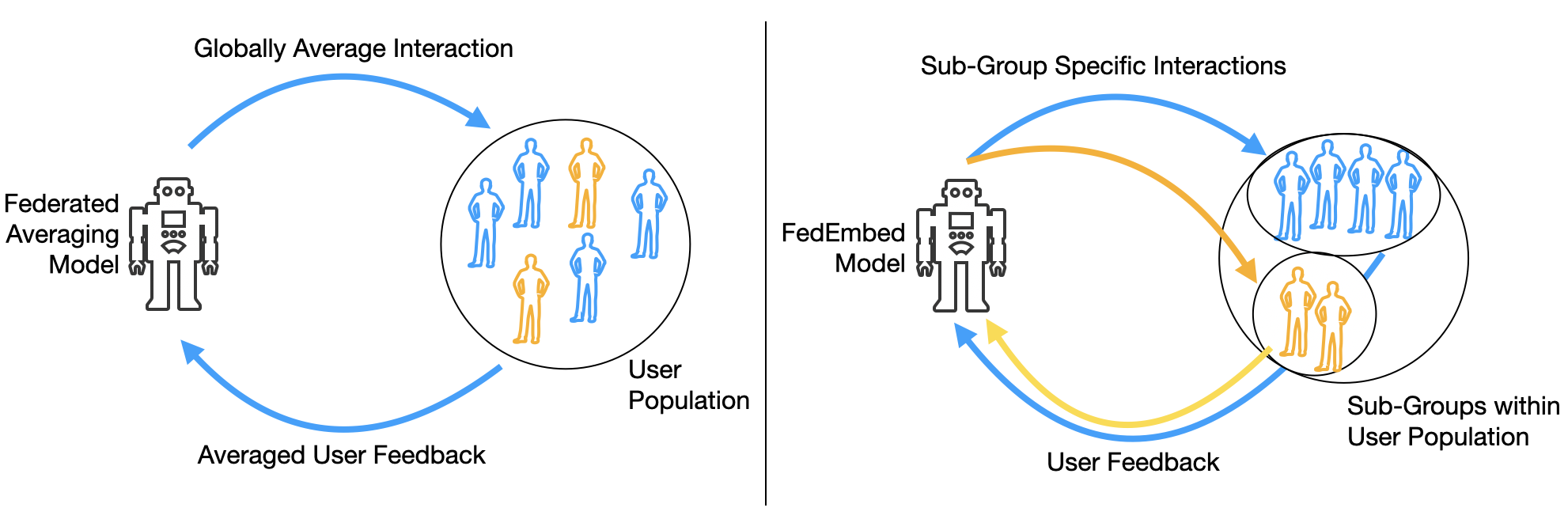}
    \caption{FedEmbed addresses the problem of using federated learning with user preferences, where the same data can have multiple labels. Federated averaging ignores individual user preferences, whereas FedEmbed learns a unique model head for each sub-population of users.}
    \label{fig:overview}
\end{figure}

However, naively applying federated averaging, or any traditional machine learning technique, will fail in domains where either data are distributed unevenly amongst the available classes, or where data contain conflicting labels. In the former case, under-represented data will not meaningfully contribute to the model since they are lost when averaging across all user gradients. This problem is compounded with the addition of differential privacy \citep{chuanxin2020federated,li2020secure}, where the shared model will either default to fit the ``average user'', or will collapse entirely. For the latter case, consider a customer service robot that interacts using speech. Some users may prefer a monotone and terse speech, whereas others may prefer an emotive voice and conversational speech~\citep{metcalf2019mirroring}. Crucially, the label associated with a given utterance is conditioned on the customer's preferences. If we naively average all customer preference labels, it is impossible to personalize a model adequately, as illustrated in Figure \ref{fig:overview}.

FedEmbed, our novel approach to personalized federated learning, works both to model user preferences, where each data sample can take conflicting labels to represent the preference of different users, and to mitigate data sparsity issues for under-represented sub-populations of users. The model is comprised of a common backbone that allows a shared feature encoder to be learned from all users, which feeds a shared representation to a separate classification head tailored for each specific sub-population. A key contribution of FedEmbed lies in learning a globally shared, sub-population independent, model using person-specific embeddings for conditioning predictions on specific users.

Our main contributions are as follows:

\begin{enumerate}
    \item FedEmbed, a novel approach to personalized, private federated learning from distributed datasets with conflicting labels.
    \item An evaluation of FedEmbed and several baselines. In the presence of differential privacy, FedEmbed allows models to personalize even as the baselines fail. We show up to 45\% improvement over baselines.
    \item A synthetic speech-preference dataset\footnote{Details for acquiring these data will be provided upon acceptance.}, presenting a challenge for future research within personalized private federated learning.
\end{enumerate}

\section{Related Work}
\label{sec:related-work}

\subsection{Personalized Federated Learning}
\label{subsec:related-work-pfl}
While the introduction of federated averaging \citep{mcmahan2017communication} enabled decentralized learning for many tasks, it has not addressed domains where data may have conflicting labels or some classes may be under-represented, which are key to personalization. Proposed solutions for such domains typically fall into one of two categories: split learning \citep{gupta2018split} or meta learning \citep{finn2017maml,nichol2018reptile}.

\textbf{Split learning} assigns a different classification head to each user, separating the learning process into representation learning and classification learning. Gradients are shared for representation learning, but each client makes local independent gradient updates to its classification head. This paradigm has been adapted for personalized federated learning: PFedMe~\citep{dinh2020pfedme}, PFedKM~\citep{tang2021pfedkm}, FedRep~\citep{collins2021fedrep}, FedMD~\citep{li2019fedmd}, and others~\citep{arivazhagan2019federated,kim2021spatio,rudovic2021personalized,paulik2021federated}. For each of these adaptations, all users contribute gradients to a shared encoder, helping the centralized model learn a robust feature representation of the data (e.g., medical images). At the same time, local classification heads learn to use the feature representation from the shared encoder to make personalized predictions (e.g., a diagnosis of a particular ailment). The personalized model head balances accurately accommodating the individuality of each user with the power of deep learning for feature extraction.

\textbf{Meta-learning} learns a function that simultaneously satisfies multiple objectives and avoids negative transfer~\citep{wang2019characterizing}. Negative transfer occurs when improvement on one task harms performance on other tasks, forcing a trade-off between objectives. Meta-learning methods, such as MAML~\citep{finn2017maml} and Reptile~\citep{nichol2018reptile}, attempt to mitigate this trade-off by considering all task objectives before making gradient updates. By considering multiple objectives and averaging out gradients, federated averaging can be viewed as a form of meta learning~\citep{jiang2019improving}. While prior work has applied meta-learning to federated learning on data with multimodal distributions~\citep{smith2017federated}, the approach does not work when there is a single task or the labels differ across users. The prior approaches learned mixtures of local and global models, where local models adapt the global model to handle individual deviations from the global mean~\citep{deng2020apfl,hanzely2020federated,hanzely2020lower}. However, in situations where user labels are directly conflicting, our results show that a global model reinforces the dominance of the average user's label, thereby not offering useful information for under-represented user models.

For real-world applications, federated learning must be extended to \textit{private} federated learning. Crucially, while private personalized learning has been applied to linear problems \cite{jain2021differentially}, no prior work has examined private personalized federated learning for deep networks. While prior work claims that split learning and federated learning approaches are privacy-preserving, \citet{rudovic2021personalized} identify that split learning masks a user's identity, but does not guarantee privacy. For guarantees of privacy, the addition of differential privacy~\citep{dwork2006calibrating,dwork2006our} to a federated learning framework is necessary. In this work, we consider the use of Gaussian differential privacy~\citep{li2020secure,chuanxin2020federated}, a widely-used approach with guaranteed bounds within which a population distribution can be accurately approximated whilst ensuring user data remain private. Our results (Section \ref{sec:experiments}) show that prior approaches to personalized federated learning cannot personalize in the presence of differential privacy.

\subsection{Personal Embeddings}
Unique personal embeddings (kept on-device) are central to the success of FedEmbed. Inspired by the use of user-specific embeddings in imitation learning~\citep{tamar2018imitation,hsiao2019learning,paleja2020interpretable}, we use personal embeddings to personalize to sub-populations of users with similar preferences. 

To cluster users with similar preferences, we make use of prototypical examples for each sub-population. Prototypes have been used for few-shot classification, where embedding networks and nearest-neighbor lookup produce accurate few-shot classifiers~\citep{snell2017prototypical}. In our work, we use nearest-neighbor lookup with personal embeddings to define clusters around a set of known prototypes.

\section{FedEmbed}
\label{sec:approach}

\begin{figure*}[t]
    \centering
    \includegraphics[width=0.9\linewidth]{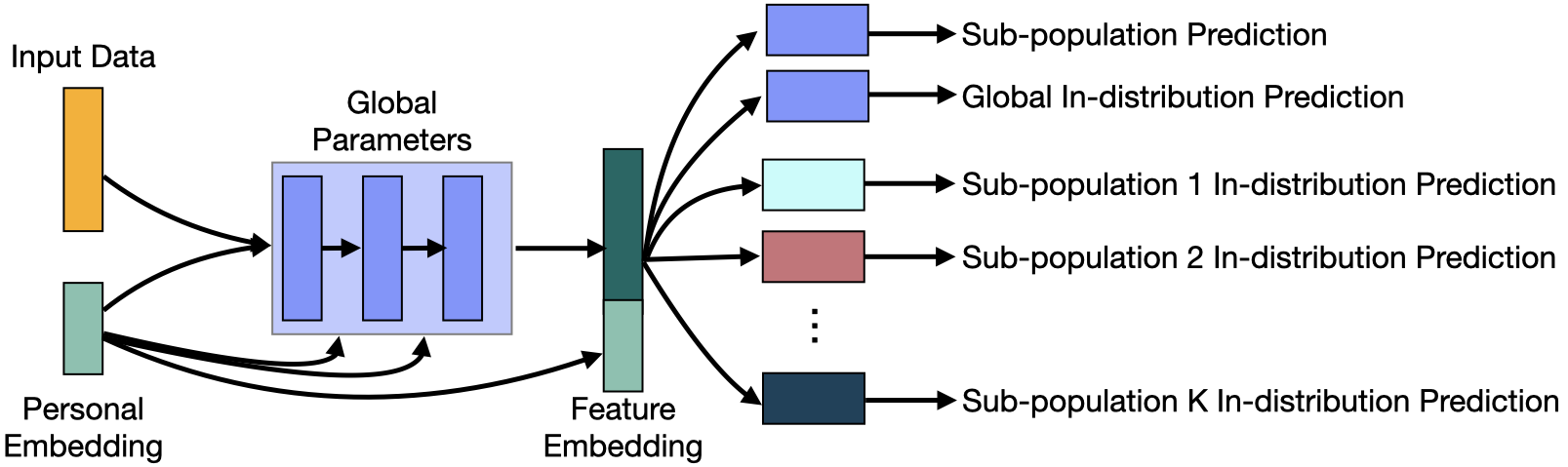}
    \caption{Overview of the FedEmbed architecture. Each client maintains a unique personal embedding on-device. The personal embedding is concatenated both to the input data and to the intermediate representations from the global parameters of a common backbone model. The output from the common backbone is a feature embedding that is passed to: 1) the client's assigned sub-population model head to classify if the input is preferred for that sub-population of users (binary classification), 2) a global model head that classifies if the input is preferred for a user, but receives gradient updates from all users (binary classification), and 3) a global sub-population prediction model-head, which predicts which of the $k$ sub-population model-heads should predict true ($k$-way classification). The purpose of the sub-population prediction head is to give structure to the feature embeddings from the common backbone (see Appendix \ref{sec:user_embeddings}), which we observe typically provides model training with a 2x speed up, and the purpose of the global in-distribution model head is to provide a latent space that is consistent across all users for learning personal embeddings.}
    \label{fig:architecture}
\end{figure*}

FedEmbed extends federated averaging to consider user-specific labels (e.g., preferences for a certain type of input data) by learning: (1) user-specific embeddings, (2) a globally shared feature encoder, (3) $k$ sub-population specific model heads that predict if a given input is in a preferred category for the respective sub-population, and (4) global model heads for predicting which of the $k$ sub-populations have a preference for the input sample and if an input sample is classified positively by a particular user. Both global heads are updated using aggregated data from all users rather than just a sub-population of similar users. The personal embeddings enable FedEmbed to cluster similar users into sub-populations that share a model head. Sub-population model heads see more data than an individual model head would, and therefore improve robustness compared to user-specific model heads. Furthermore, sub-population model heads do not receive gradients from users with conflicting labels and, therefore, offer improved personalization compared to a globally-shared model head. In this section, we present a unified overview of the learning process for FedEmbed, and then provide details on each of FedEmbed's components. Pseudocode and related training details for FedEmbed are given in Appendix \ref{sec:appendix-algorithm}.

\subsection{Learning Process}
\label{subsec:training-process}

Given a classification objective $\mathcal{L}_C$, FedEmbed learns (1) the parameters, $\theta$, of a global feature encoder, (2) the parameters, $\phi_k$, for a set of sub-population model heads $k \in [K]$, where $K$ is the number of sub-populations, (3) the parameters, $\nu$, of a global model head that predicts if an input is viewed positively or negatively for every user (regardless of their sub-population), and (4) the parameters, $\gamma$, of a global model head that predicts which of the $k$ sub-populations have a preference for the style of the given input data. 

The parameters $\gamma$ for the $k$-way classification head can be shared across all users without resulting in conflicting labels, because the type of the input data is user-independent, e.g., emotive speech is emotive regardless if it is preferred by any particular user. Furthermore, including the $k$-way prediction task embeds knowledge of the different data types within the global feature encoder. Although we find this task is not necessary for learning a shared feature encoder, it does provide structure to the shared feature representation (see Appendix \ref{sec:user_embeddings}), which regularly leads to a 2x speed up in model convergence.

When updating the model, clients pull the latest model parameters $\theta$, $\phi$, $\gamma$, and $\nu$ from the server. Each client identifies, via SOM (Section \ref{subsubsec:som-assignment}) or prototype (Section \ref{subsubsec:prototype-assignment}) assignment, a sub-population model-head. Next, clients calculate gradient updates for all parameters using their local datasets. Note that the clients maintain zero-gradients for unused model-heads, i.e., the model-heads for the sub-populations to which the user does not belong. After finishing the gradient calculation, the clients optionally add differential privacy to their computed gradients (for all parameters, including unused model heads) before sending them back to the server. In our experiments, we compare performance both with and without differential privacy.

After all clients have pushed their gradients to the server, the central server generates the updates. The $\phi_k$ gradients are averaged over all users for each of the $k$ classification heads, as the server is unaware of each client's identity or sub-population membership. The $\theta$, $\gamma$ and $\nu$ gradients are averaged over all users and applied to the shared feature encoder, and the two global model heads respectively.

\subsection{Personal Embeddings}
\label{subsec:personal-embeddings}
FedEmbed uses personal embeddings, $\vec{\epsilon}_u$, to cluster similar users. To begin, each user (i.e., client device) is assigned a personal embedding that is initialized from a uniform distribution of size $D_E$, where $D_E$ is a tuned hyper-parameter. The personal embeddings are \textbf{never} sent to the central server. At inference time on the client device, a user's personal embedding is appended to both their input samples and intermediary network representations (Figure \ref{fig:architecture}), which explicitly conditions the network predictions on relevant, user-specific details ~\citep{rownicka2019embeddings,cui2017embedding}.

At train time, following \citet{paleja2020interpretable}, a user's personal embedding receives gradient updates from all FedEmbed steps involving the user's sample-label pairs. Unlike \citet{paleja2020interpretable}, which used evidence lower-bound regularization on the gradients from the task objective function ($\mathcal{L}_C$), we found empirically that such regularization offers no benefit to the separability nor the utility of our embeddings. Additionally, because the personal embeddings stay on-device, applying privacy-preserving noise to their gradient updates is unnecessary. Therefore, the personal embeddings are updated locally without any added regularization or noise.

Embeddings learned via classification errors from different model heads may eventually occupy distinct latent spaces. However, to compare embeddings and identify meaningful user sub-populations, the embeddings must occupy a common latent space. Therefore, to obtain gradients that keep all personal embeddings in a shared latent space, we learn a globally shared classification head on the target task ($\mathcal{L}_C$). The gradients from the global classification head only update the personal embeddings, ensuring that personal embeddings are drawn from the same latent space while preventing conflicting gradients from different sub-populations from affecting the global feature encoder.

\subsection{Sub-Population Assignment}
\label{subsec:sub-group-assignment}
A sub-population represents a group of users with consistent labeling patterns (e.g., similar preferences) whose data are classified by a shared prediction head. A prior approach, PFedKM~\citep{tang2021pfedkm}, groups users via $k$-means clustering~\citep{macqueen1967some} on raw model weights and gradients. However, we observe that such clustering is noisy, impractically slow, and requires access to all user gradients simultaneously (a privacy vulnerability). Therefore, we investigated two separate approaches to sub-population assignment: (1) unsupervised clustering via self-organizing maps (SOMs)~\cite{kohonen1990self}, and (2) nearest-neighbor lookup with \emph{sub-population prototypes}.

\subsubsection{SOM Assignment}
\label{subsubsec:som-assignment}
We learn a SOM to map from personal embeddings to user sub-populations. To avoid training new model heads from scratch at every update, we reuse the model heads from the previous SOM by comparing user cluster membership in the new versus old SOMs and identify which clusters maximally overlap. For each cluster in the new SOM, the weights of the equivalent model head from the previous iteration initialize the model head associated with the new SOM cluster.

Learning the parameters of the SOM requires access to only one embedding at a time, so the SOM parameters are sent to each client. A SOM-weight update and similarity score for the client are sent back to the server, and the update from the best-scoring client is applied to each node in the SOM~\cite{kohonen1990self}. 
User privacy is preserved as personal embeddings are not aggregated on the server nor is information about original client data exposed. 

\subsubsection{Prototype Assignment}
\label{subsubsec:prototype-assignment}
While it is possible to learn sub-populations without prior knowledge via SOM assignment, the SOM can be slow to converge. If we instead assume a prototypical latent embedding exists for each known sub-population, clusters can be made denser, better-separated, and faster to converge.

We assume access to one or more prototypical users for each sub-population. Each time a user makes local gradient updates to their copy of the model, their personal embedding, $\vec{\epsilon}_u$, is first updated via a triplet loss over the prototype for the input data's current sub-population (the positive sample, $\vec{\epsilon}_p$) and a prototype for a different sub-population (the negative sample, $\vec{\epsilon}_n$). The goal is to move the personal embedding towards $\vec{\epsilon}_p$ and away from $\vec{\epsilon}_n$, where $\vec{\epsilon}_n$ is randomly sampled from all other preference prototypes. $\vec{\epsilon}_u$ is then updated according to the loss in Equation \ref{eqn:triplet_loss}, where $\alpha$ is a hyper-parameter to enforce a margin on the distance between negative samples \citep{weinberger2005distance}.
\begin{equation}
    \label{eqn:triplet_loss}
    L(\vec{\epsilon}_u, \vec{\epsilon}_p, \vec{\epsilon}_n) = (\vec{\epsilon}_u-\vec{\epsilon}_p)^2 - (\vec{\epsilon}_u-\vec{\epsilon}_n)^2 + \alpha.
\end{equation}
After the triplet-loss updates are made to $\vec{\epsilon}_u$, the user is assigned the model head corresponding to the nearest prototype embedding based on Euclidean distance. Prototype vectors are continually updated alongside the global model, keeping them in the same latent space as user embeddings. 

\subsection{Shared Feature Encoder}
\label{subsec:shared-features}
FedEmbed uses feedback from all users to learn a shared feature encoder in the form of a backbone common to all model heads. To learn this encoder, a user provides three sets of gradients to update the model by back-propogating through: 1) their sub-population model head, 2) the $k$-way classification task, and 3) the shared feature encoder.

\section{Experiments and Results}
\label{sec:experiments}
We evaluate FedEmbed on two datasets: (1) MNIST \citep{lecun1989handwritten}, and (2) a synthetic speech dataset derived from the LibriTTS corpus \citep{librispeech}. MNIST allows us to test FedEmbed on a common federated learning test-bed with clear and discrete sub-populations (i.e., digits) \cite{mcmahan2017communication,dinh2020pfedme,tang2021pfedkm}. Our synthetic speech dataset consists of 12,420 utterances from 20 synthetic voices created by interpolating the parameterized form of voices from eight human speakers using the method in \citet{speech_generator}. All speech samples are transformed into spectrograms. Interpolated voices present a unique challenge, because the sub-populations are less distinct than is the case for MNIST digits. Additionally, the dimensionality of the spectrograms used as input to the networks is larger than that of MNIST images --- we trim or pad all spectrograms to be 200x375 compared to 28x28 images for MNIST.

For each dataset, we construct a set of simulated users according to a set of rules (see Sections \ref{sec:expMNIST} and \ref{sec:expSpeech} for MNIST and speech experiments respectively). Each user has a local dataset, $D_u$, of uniform size, and is assigned to a sub-population $k \in [K]$. For MNIST, the user sub-populations correspond to digits, whereas for our speech dataset user sub-populations correspond to a particular interpolated voice. For each sample in $D_u$, if the digit/voice for the sample matches the assigned sub-population preference, $k$, then the sample is labeled positive, otherwise the sample is labeled negative. In our experiments, each user has 30 samples ($D_u$), 20 of which are used for training and 10 of which are used for testing. The data for each user are split evenly between positive and negative labels.

In our experiments, we consider single-channel image data. To concatenate a user embedding, $\vec{\epsilon}_u$, of size $D_E \times 1$ with the input data, we concatenate
a given input image, $X$, with a diagonal matrix, $I\vec{\epsilon}_u$, where $I$ is the identity matrix, forming a multi-channel image. This process enables gradient information to back-propagate to the original embedding. 

A centralized model without specialized model heads nor personal embeddings is incapable of fitting the preference datasets, because the label for a sample is conditioned on the user. There is no way to predict the label without information about the user, which could come from a sub-population model head or from a preference embedding.

For each dataset, we evaluate FedEmbed using average F1 score across all $K$ sub-populations in both a class balanced and class imbalanced setting. For the balanced data, all sub-populations are equally represented; for the imbalanced data, we select two sub-populations to be over-represented, while the others are under-represented. The class imbalanced experiments evaluate the robustness of the shared feature encoder and determine whether different approaches to personalized federated learning are well-suited to class imbalanced problems. Finally, we experiment with adding Gaussian differential privacy to all of our experiments \citep{chuanxin2020federated,li2020secure}, representing, to the best of our knowledge, the first personalized federated learning experiments to feature differential privacy. For all experimental settings we compare FedEmbed, FedEmbed derivatives, and all of our baseline approaches.

\subsection{Baselines}
\label{subsec:baselines}
We compare FedEmbed to standard federated averaging, three personalized federated learning frameworks from prior work, and ablations of FedEmbed. Specifically, we compare the approaches outlined in Table~\ref{tab:approaches}.
\begin{table*}[t]
\caption{The approaches to personalized private federated learning considered in our work.}
\label{tab:approaches}
\begin{tabular}{|r|p{12cm}|}
\hline
\textbf{Global}&standard federated averaging, with a single model head \citep{mcmahan2017communication}\\\hline
\textbf{Global+}&federated averaging with personal embeddings for each user\\\hline
\textbf{PFedM}&a shared feature encoder with \emph{personal} model heads regularized to minimize parameter difference from a global model head \citep{dinh2020pfedme}\\\hline
\textbf{FedRep}&a shared feature encoder with \emph{personal} model heads and no regularization \citep{collins2021fedrep}\\\hline
\textbf{PFedKM}&a shared feature encoder with sub-population model heads assigned by $K$-means clustering over user gradients \citep{tang2021pfedkm} \footnote{Empirically, we found SOM clustering to be more accurate than K-Means, even for PFedKM's gradient clustering , and so our implementation of PFedKM uses SOM for clustering to provide a fair comparison.}\\\hline
\textbf{FedEmbed-SOM}&FedEmbed with SOM sub-population assignment\\\hline
\textbf{FedEmbed-Personal}&FedEmbed with each user assigned to their own personal cluster\\\hline
\textbf{FedEmbed-Prototype}&FedEmbed with prototype sub-population assignment\\\hline
\textbf{FedEmbed-Type}&FedEmbed with the ground-truth preference sub-population for each user\\\hline
\end{tabular}
\end{table*}

FedEmbed-Type provides an estimate on the upper-bound on the performance of FedEmbed, as this removes clustering errors from our approach as we assume the sub-population of each user is known and we assign the user the corresponding model head. FedEmbed-Personal is similar to PFedMe and FedRep, but it includes the addition of personal embeddings for users.

\subsection{MNIST Experiments}\label{sec:expMNIST}
Our MNIST experiments feature 10 sub-populations, one for each unique digit (i.e., a user will label a single digit type positively and all others negatively). We use all 60,000 training MNIST images for a total of 900 users per sub-population, reflecting a real-world federated learning deployment in which each user has only a few labeled samples (making it too small to learn an individual model), but the full dataset is large enough to learn a useful model.

For our balanced-data experiments, all 10 sub-populations are equally represented. For our imbalanced dataset, type 0 represents 25\% of users, type 1 represents 15\% of users, types 2--5 are 10\% each, and types 6--9 are 5\% each, enabling us to examine performance on a clear majority class, and a set of larger and smaller minority classes.

\subsubsection{MNIST Results}

\begin{figure*}[t]
    \centering
    \includegraphics[width=\linewidth]{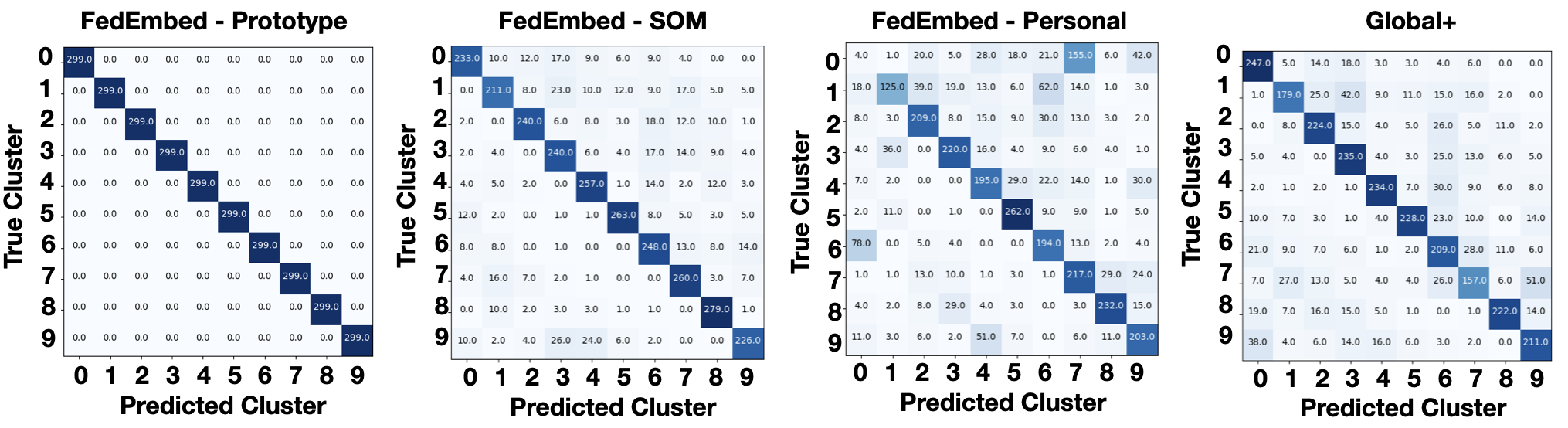}
    \caption{Confusion matrices for ground-truth user preferences and predicted user-preferences using personal embeddings derived from our balanced MNIST dataset. FedEmbed-Prototype perfectly recovers user preferences, and all approaches are able to separate most preferences.}
    \label{fig:mnist-plot}
\end{figure*}

We present final average F1 scores for all four experiments in Table \ref{tab:mnist-results}, and preference-assignment confusion matrices in Figure \ref{fig:mnist-plot}. As expected, FedEmbed-Type presents an upper-bound and exceeds all baselines for both balanced and imbalanced data, with and without differential privacy, achieving between 12\% and 41\% gains relative to prior work. The performance of all personalization approaches using personal model heads (FedRep, PFedMe, and FedEmbed-Personal) severely deteriorates when differential privacy is added due to insufficient gradient information to overcome the privacy-preserving noise. Finally, we note that Global, Global+, PFedKM, and FedEmbed-SOM appear to improve with imbalanced data relative to balanced data. However, closer examination of their class-wise performance (Table \ref{tab:mnist-imba}) shows these methods have fit the most common preference type (i.e., the 25\% majority) at the expense of all other preference types.

\begin{table*}[th]
\centering
\caption{Mean and standard deviation F1 scores over five independent trails with MNIST data (bold indicates best)}
\label{tab:mnist-results}
\begin{tabular}{c|cc|cc}
Method & \multicolumn{2}{c|}{Balanced Data} & \multicolumn{2}{c}{Imbalanced Data} \\\hline
Privacy & None & Gaussian & None & Gaussian \\\hline\hline
FedRep & 0.78 $\pm$ 0.02 & 0.51 $\pm$ 0.02 & 0.78 $\pm$ 0.02 & 0.53 $\pm$ 0.01 \\
PFedMe & 0.76 $\pm$ 0.04 & 0.50 $\pm$ 0.02 & 0.76 $\pm$ 0.05 & 0.50 $\pm$ 0.04 \\
PFedKM & 0.48 $\pm$ 0.00 & 0.49 $\pm$ 0.00 & 0.55 $\pm$ 0.02 & 0.48 $\pm$ 0.03 \\
Global & 0.40 $\pm$ 0.02 & 0.38 $\pm$ 0.06 & 0.55 $\pm$ 0.06 & 0.38 $\pm$ 0.06 \\
Global+ & 0.39 $\pm$ 0.06 & 0.36 $\pm$ 0.04 & 0.54 $\pm$ 0.08 & 0.34 $\pm$ 0.00 \\
FedEmbed-Personal & 0.76 $\pm$ 0.01 & 0.52 $\pm$ 0.01 & 0.73 $\pm$ 0.02 & 0.52 $\pm$ 0.02 \\
FedEmbed-SOM & 0.60 $\pm$ 0.05 & 0.47 $\pm$ 0.07 & 0.69 $\pm$ 0.02 & 0.53 $\pm$ 0.02 \\
\textbf{FedEmbed-Prototype} & \textbf{0.88 $\pm$ 0.02} & 0.69 $\pm$ 0.04 & 0.87 $\pm$ 0.02 & 0.69 $\pm$ 0.05 \\
\textbf{FedEmbed-Type} & \textbf{0.88 $\pm$ 0.01} & \textbf{0.72 $\pm$ 0.05} & \textbf{0.88 $\pm$ 0.03} & \textbf{0.77 $\pm$ 0.05}
\end{tabular}
\end{table*}

\subsection{Speaker Preference Experiments}\label{sec:expSpeech}
Our synthetic speech experiments use a population of 900 split into 20 sub-populations, one for each synthetic voice (i.e., a user will label one voice positively and all others negatively, regardless of the spoken words). This presents a significantly more challenging problem than our MNIST experiment. The objective is to assess a scenario where user preferences are not necessarily distinct --- in our case, two different voice-types might be two different mixtures of the same source voices.

For our balanced dataset experiments, all 20 sub-populations are represented with the same number of users. For our imbalanced dataset, voice 0 is preferred by 20\% of the users, voice 5 is preferred by 8\% of the users, and all other voices are preferred 4\% of the time, again presenting a comparison between one clear majority class, a larger minority, and a long tail of minority classes.

\subsubsection{Speaker Preference Results}

Table \ref{tab:speech-results} shows mean and standard deviation F1 scores for our four speaker-preference experiments (balanced/imbalanced, with/without differential privacy), and preference-assignment confusion matrics are given in Figure \ref{fig:speaker-cluster-plot}. Across the board, performance on the speaker-preference data is lower than on the MNIST dataset, signaling the significant increase in dataset-difficulty. As in the MNIST experiments, we observe FedEmbed-Type and FedEmbed-Prototype exceed all other personalized federated learning approaches both with and without differential privacy, achieving between 15\% and 35\% improvements relative to prior work. We again observe that Gaussian differential privacy adversely affects performance, with the F1 score for all baselines dropping below 0.5. As in the MNIST experiments, only approaches with effective grouping of similar users (i.e., FedEmbed-Prototype and FedEmbed-Type) overcome the noise of differential privacy to learn a useful model that personalizes. Finally, we again note an apparent increase in performance when the dataset is imbalanced. As in the MNIST results, we observe that this can be attributed to the models fitting the most populous classes at the expense of the under-represented preferences (Table \ref{tab:speaker-imba}).

\begin{table*}[th]
\centering
\caption{Mean and standard deviation F1 scores for five independent runs with speech data (bold indicates best)}
\label{tab:speech-results}
\begin{tabular}{c|cc|cc}
Method & \multicolumn{2}{c|}{Balanced Data} & \multicolumn{2}{c}{Imbalanced Data} \\\hline
Privacy & None & Gaussian & None & Gaussian \\
\hline\hline
FedRep & 0.52 $\pm$ 0.09 & 0.49 $\pm$ 0.00 & 0.65 $\pm$ 0.10 & 0.49 $\pm$ 0.00 \\
PFedMe & 0.57 $\pm$ 0.05 & 0.49 $\pm$ 0.01 & 0.57 $\pm$ 0.05 & 0.49 $\pm$ 0.03 \\
PFedKM & 0.49 $\pm$ 0.02 & 0.49 $\pm$ 0.01 & 0.49 $\pm$ 0.01 & 0.49 $\pm$ 0.00 \\
Global & 0.34 $\pm$ 0.01 & 0.38 $\pm$ 0.06 & 0.44 $\pm$ 0.05 & 0.34 $\pm$ 0.02 \\
Global+ & 0.35 $\pm$ 0.01 & 0.33 $\pm$ 0.00 & 0.42 $\pm$ 0.12 & 0.33 $\pm$ 0.00 \\
FedEmbed-Personal & 0.63 $\pm$ 0.02 & 0.44 $\pm$ 0.05 & 0.68 $\pm$ 0.01 & 0.49 $\pm$ 0.00 \\
FedEmbed-SOM & 0.51 $\pm$ 0.02 & 0.47 $\pm$ 0.07 & 0.54 $\pm$ 0.03 & 0.47 $\pm$ 0.02 \\
\textbf{FedEmbed-Prototype} & \textbf{0.78 $\pm$ 0.01} & \textbf{0.61 $\pm$ 0.06} & 0.63 $\pm$ 0.07 & 0.53 $\pm$ 0.06 \\
\textbf{FedEmbed-Type} & \textbf{0.78 $\pm$ 0.05} & 0.60 $\pm$ 0.08 & \textbf{0.76 $\pm$ 0.00} & \textbf{0.56 $\pm$ 0.08}
\end{tabular}
\end{table*}

\section{Discussion}
Comparing with existing work, we observe that FedEmbed offers improvements over the state of the art in private federated learning when personalizing models. While previous approaches, such as PFedMe and FedRep, offer useful advances for personalized federated learning, no prior approach offers accurate personalization with differential privacy. By enabling private personalized federated learning, FedEmbed represents an advancement for private and personalized federated learning.

FedEmbed further improves performance for under-represented sub-populations relative to all baseline approaches. While a global approach may appear successful on over-represented classes, existing approaches reduce performance for under-represented classes. Our results suggest that fairness (i.e., equitable performance) in federated learning has a long way to go, but that intelligently grouping related users can, in part, mitigate this problem and provide improved performance for under-represented classes (Table \ref{tab:mnist-imba}) even with a shared feature encoder. 



By ablating FedEmbed's component contributions, we see how FedEmbed compares to baselines and identify the components of FedEmbed that contribute most to its performance. As noted above, FedEmbed-Personal is similar to FedRep and PFedMe, while Global+ is similar to Global with the addition of personal embeddings. Across these comparisons, we see that personal embeddings on their own do not make a difference on the simpler MNIST task as performance is comparable for methods with and without personal embeddings. On the more challenging speaker-preference task, we observe a more pronounced difference between methods with and without personal embeddings, with FedEmbed-Personal outperforming baselines by nearly 20\% on the balanced dataset without differential privacy. This result suggests that the personal embeddings in FedEmbed provide performance benefits for more challenging problems, but on simpler problems the benefits can be attributed to meaningful sub-group assignment by clustering personal embeddings.

The clustering results from our MNIST dataset (Figure \ref{fig:mnist-plot}) suggest that all embedding-based approaches (i.e., Global+, FedEmbed-Personal, FedEmbed-SOM, and FedEmbed-Prototype) should discover similar user sub-groups and have the same ability to accurately cluster users to the appropriate sub-groups. This result suggests that an approach like Global+ can achieve similar performance to FedEmbed, if the global model learns to condition on the personal embedding. However, we observe that Global+ performs significantly worse than FedEmbed approaches.

Results from our speaker-preference dataset (Figure \ref{fig:speaker-cluster-plot}) help to shed light on the discrepancy in performance between embedding-based approaches. The speaker-preference results suggest that less-structured approaches (i.e. Global+, FedEmbed-SOM, and FedEmbed-Personal) are significantly worse at accurately clustering users. This disparity in cluster assignment may help to explain the disparity in mean F1 performance between FedEmbed-Prototype and other techniques (Tables \ref{tab:mnist-results} and \ref{tab:speech-results}). Techniques with less prior knowledge may assign users to the wrong sub-populations and therefore train model-heads on conflicting preference labels. For example, Global+ and FedEmbed-SOM have no prototypical user information, potentially leading these approaches to form inaccurate clusters and impede the learning process. This result may have only appeared in the speaker-preference task because the speech data has much higher dimensionality and represents twice as many classes as MNIST.

Considering SOM- and prototype-assignment to ground-truth types, we see the importance of correctly clustering users. Ground-truth assignment is almost always the best, while naive clustering with SOM is consistently the least reliable approach, having little prior knowledge to solve the unsupervised learning problem. Despite the under-performance of FedEmbed-SOM, we find FedEmbed-SOM far exceeds PFedKM both in performance and runtime. Because PFedKM clusters gradients for an entire federated learning model, the approach is extremely slow, memory-demanding, and computationally intensive. FedEmbed-SOM is faster and performs better than PFedKM, as personal embeddings are easier to cluster than entire models. Finally, we observe that prototype assignment (FedEmbed-Prototype) works well, often performance is close to the ground-truth type assignment (FedEmbed-Type). 

\section{Limitations and Future Work}
While FedEmbed represents a significant step forward for private and personalized federated learning, there are limitations that must be addressed in future work. First, we assume knowledge of at least the number of sub-populations within our global population (for FedEmbed-SOM), or access to a prototypical member of each sub-population (for FedEmbed-Prototype). While such data could be captured through a small user study, observation, or discussions with domain experts, it does represent an area for improvement. In follow-up experiments, we determined that overestimating the number of clusters (e.g., looking for 20 types of MNIST users instead of 10) did not adversely affect performance, but more study is needed in this area.

Second, we observe that the performance of FedEmbed deteriorates as class representation declines and Gaussian privacy is enabled. While FedEmbed preserves under-represented class performance better than alternative methods, there is room to improve for under-represented sub-populations. We believe FedEmbed can offer a path to improved fairness in federated learning, though such improvements must be carefully monitored to ensure that overall performance gains are reflected in improvements across all classes. Our results show that naive SOM-assignment does not work as well as prototype-assignment for imbalanced data, indicating that domain-knowledge about the distribution of users will yield improved results.

Finally, FedEmbed inherently assigns users to sub-populations that will maximize the number of positive predictions for their preferred type. This assignment could present problems with users who change preferences over time, or for whom a sub-population does not yet exist (e.g., the user does not like monotone or emotive speech, but the robot expects only those two preference types). Future work will study ways to improve personalization over time, techiques to capture changes in preferences over time, and methods to discover new sub-populations over time.

\section{Conclusions}
We have presented FedEmbed, a new approach to personalized federated learning by clustering related users into sub-populations that share a model head, rather than assigning each user their own unique model head or averaging over all preferences. We showed that such clustering enables more robust preference learning for balanced and imbalanced datasets, outperforming baselines when considering under-represented sub-populations. We showed that FedEmbed learns preferences when applying differential privacy, unlike our baselines. We presented an ablation of FedEmbed examining the effects of different clustering approaches and showing that prototype-assignment perfectly recovers user preferences. Finally, we outlined areas for future work to improve personalized federated learning, such as removing clustering assumptions or improved preference learning over time.

\bibliography{bib}
\bibliographystyle{report}

\onecolumn
\appendix
\section{FedEmbed Algorithm}
\label{sec:appendix-algorithm}
To more clearly explain the learning and update procedures for FedEmbed, in Algorithm \ref{alg:training_loop} we present an algorithmic overview of the update process. Our training process is repeated here from the main paper, but now annotated by lines in Algorithm \ref{alg:training_loop} for improved clarity. 

At the beginning of each training update, clients pull down the latest set of model parameters $\theta$ and $\phi$ from the server (Algorithm \ref{alg:training_loop}: line 6). Each client $u \in [U]$ first identifies which model-head to use for inference and training (Algorithm \ref{alg:training_loop}: lines 8--14). This model-head selection can be accomplished through the SOM-assignment process or prototype assignment. Once a client has selected a model head, the client calculates gradient updates for $\theta$ and their model head, $\phi$, using their local dataset, $D_u$ (Algorithm \ref{alg:training_loop}: lines 15--21). Note that the client maintains zero-gradients for model-heads they do not use. After finishing the gradient calculation, the client can optionally add differential privacy to their computed gradients before sending them back to the server.

After all clients have computed gradients and pushed them to the server, the central server averages $\theta$ and $\phi$ gradients, again optionally adding differential privacy to the result (Algorithm \ref{alg:training_loop}: lines 24--26). The server then applies the gradient steps to the shared feature encoder parameters and the model-head parameters (Algorithm \ref{alg:training_loop}: lines 27--2), and the process repeats.

\begin{algorithm*}
\caption{FedEmbed Training Procedure}
\label{alg:training_loop}
\begin{algorithmic}[1]
\STATE {\bfseries Given:} Classification objective, $\mathcal{L}_C$, Mode-prediction objective, $\mathcal{L}_M$, Learning rate $\alpha$
\STATE {\bfseries Server Initializes:} Global encoder, $\theta$, $\phi_k$ model heads,
\STATE {\bfseries Clustering Initialization:} SOM $\Psi$ or Prototype Embeddings $\mathcal{E}$
\STATE {\bfseries Client Initializes:} Personal embedding, $\vec{\epsilon}_u$
\WHILE{Training:}
\STATE Clients pull down $\theta$ and $k$ model heads
\\ \hrulefill \\
{\bfseries Client Update} 
\\ \hrulefill
\STATE {\bfseries Given: } Local user dataset, $D_u$
\IF{Prototype-assignment}
\STATE {Identify $\vec{\epsilon}_p$ from $D_u$, sample $\vec{\epsilon}_n$ from $\mathcal{E}$}
\STATE{$\vec{\epsilon}_u \leftarrow \vec{\epsilon}_u + \nabla_{\vec{\epsilon}_u} L(\vec{\epsilon}_u, \vec{\epsilon}_p, \vec{\epsilon}_n)$}
\STATE {Model head, $\phi$, assigned as nearest neighbor to $\vec{\epsilon}_u$ in $\mathcal{E}$}
\ELSE
\STATE {Model head, $\phi$, assigned by $\Psi (\vec{\epsilon}_u) $}
\ENDIF
\STATE $\nabla_{u_\theta} = 0$
\STATE $\nabla_{u_{\phi}} = 0$
\FOR{$\vec{x}, y, m$ in $D_u$}
\STATE $\nabla_{u_\theta} =  \nabla_{u_\theta} + \nabla_\theta\mathcal{L}_C(\phi(\theta(\vec{x})), y) + \nabla_\theta\mathcal{L}_M(\theta(\vec{x}, m)$
\STATE  $\nabla_{u_{\phi}} = \nabla_{u_{\phi}} + \nabla_{\phi}\mathcal{L}_C(\phi(\theta(\vec{x})), y)$
\STATE $\vec{\epsilon}_u \leftarrow \vec{\epsilon}_u + \nabla_{\vec{\epsilon}_u}\mathcal{L}_C(\phi(\theta(\vec{x})), y)$
\ENDFOR
\STATE {Optionally, add differential privacy to $\nabla_{u_\theta}$ and $\nabla_{u_{\phi}}$}
\\
 \hrulefill
\\ {\bfseries Server Update}
\\ \hrulefill
\STATE {\bfseries Given: } Gradient updates $\nabla_{u_\theta}$ and  $\nabla_{u_\phi}$ from active users $u \in [U]$
\STATE {$\nabla_\theta = \frac{1}{U}\sum_{u=0}^{U}\nabla_{u_\theta}$}
\STATE {$\nabla_\phi = \frac{1}{U}\sum_{u=0}^{U}\nabla_{u_\phi}$}
\STATE {Optionally, add differential privacy to $\nabla_\theta$ and $\nabla_\phi$}
\STATE {$\theta \leftarrow \theta + \alpha \nabla_\theta$}
\STATE {$\phi \leftarrow \phi + \alpha \nabla_\phi$}
\ENDWHILE
\end{algorithmic}
\end{algorithm*}

\newpage

\section{Architecture Descriptions}\label{sec:architectures}

\subsection{MNIST Models}

The architectures for the MNIST experiments are as follows (following the PyTorch API), where an input image $x$ has dimensions $x \in \mathbb{R}^{(28 \times 28 \times 2)}$:

\begin{minipage}[t]{0.55\textwidth}
\begin{lstlisting}[language=Python,numbers=none]
# feature encoder shared by all 
# sub-population prediction heads
global_encoder = Sequential(
     Conv2d(2, 16, 3, stride=2, padding=1),
     LayerNorm([16, 14, 14]),
     Conv2d(16, 32, 3, stride=2, padding=1), 
     LayerNorm([32, 7, 7]), 
     Conv2d(32, 64, 7, stride=1, padding=0), 
     LayerNorm([64, 1, 1])
  )

\end{lstlisting}
\end{minipage}
\begin{minipage}[t]{0.45\textwidth}
\begin{lstlisting}[language=Python,numbers=none]
# the global head used to predict
# preferences for all users and 
# sub-populations
global_prediction_head = Linear(92, 2)

# the prediction heads for each 
# sub-population
model_head_0 = Linear(92, 2)
...
model_head_9 = Linear(92, 2)

# the k-way classifier that predicts
# group ID
classifier_head = Linear(92, 10)
\end{lstlisting}
\end{minipage}

The local optimizer is Adam with PyTorch's default parameters, and the central optimizer is Adam with $lr=1$. The value of $\alpha$ for the triplet margin loss (Equation \ref{eqn:triplet_loss}) is $1.0$. The size of the personal embedding, $\vec{\epsilon}_u$, is $28 \times 1$. 

\subsection{Speech Models}

The architectures for the speech experiments are as follows (following the PyTorch API), where an input spectrogram $x$ has dimensions $x \in \mathbb{R}^{(200 \times 375 \times 2)}$:

\begin{minipage}[t]{0.55\textwidth}
\begin{lstlisting}[language=Python,numbers=none]
# feature encoder shared by all 
# sub-population prediction heads
global_encoder = Sequential(
     Conv2d(2, 16, kernel=(3, 5), stride=(2, 3)), 
     AvgPool2d((2, 3)), 
     LayerNorm([16, 50, 41]), 
     Conv2d(16, 32, kernel=(5, 5), stride=(2, 2)), 
     AvgPool2d(3, 3)), 
     LayerNorm([32, 7, 6]), 
     Conv2d(32, 64, kernel=(3, 2), stride=(2, 2)), 
     LayerNorm([64, 3, 3]), 
     Flatten(1), 
     Linear(576, 256), 
     LayerNorm([256])
  )
\end{lstlisting}
\end{minipage}
\begin{minipage}[t]{0.45\textwidth}
\begin{lstlisting}[language=Python,numbers=none]
# the global head used to predict 
# preferences for all users and  
# sub-populations
global_prediction_head = Linear(92, 2)

# the prediction heads for each 
# sub-population
model_head_0 = Linear(92, 2)
...
model_head_20 = Linear(92, 2)

# the k-way classifier that predicts 
# group ID
classifier_head = Linear(92, 20)
\end{lstlisting}
\end{minipage}

The local optimizer is Adam with PyTorch's default parameters, and the central optimizer is Adam with $lr=1$. The value of $\alpha$ for the triplet margin loss (Equation \ref{eqn:triplet_loss}) is $1.0$. The size of the personal embedding, $\vec{\epsilon}_u$, is $200 \times 1$. 

\newpage

\section{Effect of $k$-way classification on user embeddings structure}\label{sec:user_embeddings}

\begin{figure*}[h]
\centering
\begin{subfigure}{0.47\textwidth}
        \includegraphics[width=\textwidth]{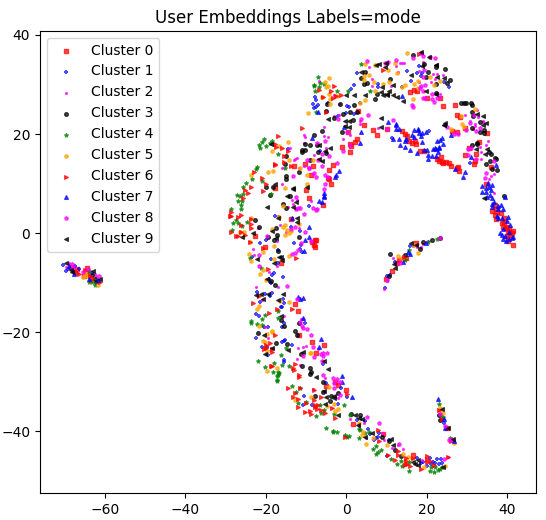}
        \caption{User embeddings from a model without $k$-way classification}
    \end{subfigure}
    ~~
    \begin{subfigure}{0.47\textwidth}
        \includegraphics[width=\textwidth]{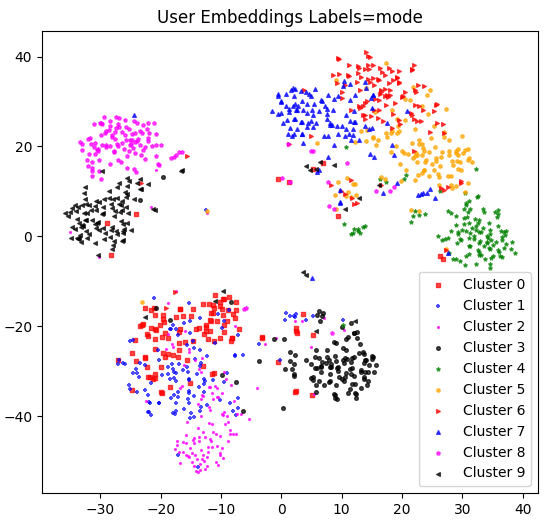}
        \caption{User embeddings from a model with $k$-way classification}
    \end{subfigure}
\caption{A TSNE projection of the user embeddings from an MNIST model both: (a) without the $k$-way classification head, and (b) with the $k$-way classification head.}
\label{fig:user_embeddings}
\end{figure*}

Figure \ref{fig:user_embeddings} visualizes the impact of the $k$-way classification head on how the learned user embeddings cluster in a 2D space (projected via TSNE). Without the use of the $k$-way classification head (Figure \ref{fig:user_embeddings}(a)), the user embeddings do not separate by sub-population. However, with the use of the $k$-way classification head, we can see the user embeddings separating by sub-population \ref{fig:user_embeddings}(b)). These figures illustrate the information about the user types that the k-way classification head transfers into the user embeddings.

\newpage

\section{Class-wise Preference Performance on Imbalanced Data}

\subsection{MNIST}

The mean and standard deviation F1-score per federated learning approach (with and without Gaussian privacy) broken down by the prevalence of a user preference type in the training dataset are reported in Table \ref{tab:mnist-imba}. For the imbalanced dataset scenario, 25\% of the dataset consists of users with a preference for digit 0, 15\% of users with a preference for digit 1, 10\% each with a preference for digits 2 - 5, and 5\% each with a preference for digits 6-9.

\begin{table*}[th]
\centering
\caption{Mean F1 scores over five independent trails for imbalanced MNIST data (bold indicates best)}
\label{tab:mnist-imba}
\begin{tabular}{c|llll}
\hline
Method & \multicolumn{4}{c}{No Privacy}  \\\hline
Sub-population representation & \multicolumn{1}{c}{25\%} & \multicolumn{1}{c}{15\%} & \multicolumn{1}{c}{10\%} & \multicolumn{1}{c}{5\%} \\
\hline\hline
PFedMe & 0.77 $\pm$ 0.19 & 0.79 $\pm$ 0.17 & 0.74 $\pm$ 0.14 & 0.76 $\pm$ 0.15 \\
PFedKM & 0.66 $\pm$ 0.04 & 0.57 $\pm$ 0.04 & 0.49 $\pm$ 0.03 & 0.43 $\pm$ 0.03 \\
FedRep & 0.80 $\pm$ 0.18 & 0.79 $\pm$ 0.15 & 0.75 $\pm$ 0.14 & 0.77 $\pm$ 0.15 \\
Global & 0.82 $\pm$ 0.10 & 0.43 $\pm$ 0.09 & 0.38 $\pm$ 0.03 & 0.36 $\pm$ 0.02 \\
Global+ & 0.83 $\pm$ 0.10 & 0.40 $\pm$ 0.16 & 0.41 $\pm$ 0.08 & 0.36 $\pm$ 0.10 \\
FedEmbed-Personal & 0.72 $\pm$ 0.16 & 0.75 $\pm$ 0.13 & 0.74 $\pm$ 0.12 & 0.74 $\pm$ 0.15 \\
FedEmbed-SOM & 0.91 $\pm$ 0.14 & 0.69 $\pm$ 0.08 & 0.61 $\pm$ 0.13 & 0.45 $\pm$ 0.14 \\
\textbf{FedEmbed-Prototype} & 0.82 $\pm$ 0.16 & 0.86 $\pm$ 0.06 & \textbf{0.85 $\pm$ 0.07} & \textbf{0.89 $\pm$ 0.05} \\
\textbf{FedEmbed-Type} & \textbf{0.92 $\pm$ 0.27} & \textbf{0.91 $\pm$ 0.26} & 0.84 $\pm$ 0.23 & 0.83 $\pm$ 0.24 \\ \hline
& \multicolumn{4}{c}{Gaussian Privacy} \\\hline \hline
PFedMe & 0.49 $\pm$ 0.06 & 0.50 $\pm$ 0.07 & 0.49 $\pm$ 0.05 & 0.48 $\pm$ 0.05 \\
PFedKM & 0.49 $\pm$ 0.02 & 0.47 $\pm$ 0.03 & 0.48 $\pm$ 0.02 & 0.48 $\pm$ 0.02 \\
FedRep & 0.55 $\pm$ 0.05 & 0.49 $\pm$ 0.05 & 0.51 $\pm$ 0.03 & 0.50 $\pm$ 0.03 \\
Global & 0.39 $\pm$ 0.10 & 0.37 $\pm$ 0.10 & 0.38 $\pm$ 0.09 & 0.37 $\pm$ 0.08 \\
Global+ & 0.33 $\pm$ 0.10 & 0.34 $\pm$ 0.08 & 0.34 $\pm$ 0.07 & 0.34 $\pm$ 0.07 \\
FedEmbed-Personal & 0.51 $\pm$ 0.06 & 0.49 $\pm$ 0.05 & 0.50 $\pm$ 0.03 & 0.50 $\pm$ 0.04 \\
FedEmbed-SOM & 0.65 $\pm$ 0.13 & 0.49 $\pm$ 0.13 & 0.49 $\pm$ 0.08 & 0.45 $\pm$ 0.07 \\
\textbf{FedEmbed-Prototype} & \textbf{0.85 $\pm$ 0.13} & 0.59 $\pm$ 0.14 & 0.69 $\pm$ 0.18 & \textbf{0.69 $\pm$ 0.16} \\
\textbf{FedEmbed-Type} & 0.79 $\pm$ 0.20 & \textbf{0.90 $\pm$ 0.22} & \textbf{0.72 $\pm$ 0.19} & 0.67 $\pm$ 0.19
\end{tabular}
\end{table*}

\newpage

\subsection{Speaker Preference}
The mean and standard deviation F1-score per federated learning approach (with and without Gaussian privacy) broken down by the prevalence of a user preference type in the dataset are presented in Table \ref{tab:speaker-imba}. For the imbalanced speaker preference data scenario, 20\% of the users prefer voice 0, 8\% prefer voice 5, and 4\% each for the remaining 18 voices.

\begin{table*}[th]
\centering
\caption{Mean F1 scores over five independent trails for imbalanced Speaker preference data (bold indicates best)}
\label{tab:speaker-imba}
\begin{tabular}{c|lll}
\hline
Method                      & \multicolumn{3}{c}{No Privacy}  \\\hline
Sub-population representation       & \multicolumn{1}{c}{20\%} & \multicolumn{1}{c}{8\%} & \multicolumn{1}{c}{4\%}         \\
\hline\hline
PFedMe & 0.66 $\pm$ 0.11 & 0.62 $\pm$ 0.10 & 0.54 $\pm$ 0.06 \\
PFedKM & 0.49 $\pm$ 0.01 & 0.49 $\pm$ 0.02 & 0.48 $\pm$ 0.02 \\
FedRep & 0.65 $\pm$ 0.15 & 0.52 $\pm$ 0.08 & 0.55 $\pm$ 0.09 \\
Global & 0.57 $\pm$ 0.24 & 0.49 $\pm$ 0.16 & 0.38 $\pm$ 0.11 \\
Global+ & 0.51 $\pm$ 0.22 & 0.53 $\pm$ 0.20 & 0.38 $\pm$ 0.12 \\
FedEmbed-Personal & 0.78 $\pm$ 0.01 & 0.68 $\pm$ 0.08 & 0.65 $\pm$ 0.07 \\
FedEmbed-SOM & 0.65 $\pm$ 0.13 & 0.60 $\pm$ 0.09 & 0.50 $\pm$ 0.08 \\
\textbf{FedEmbed-Prototype} & \textbf{0.79 $\pm$ 0.09} & 0.71 $\pm$ 0.20 & 0.54 $\pm$ 0.18 \\
\textbf{FedEmbed-Type} & 0.73 $\pm$ 0.05 & \textbf{0.92 $\pm$ 0.04} & \textbf{0.73 $\pm$ 0.13} \\ \hline            
                      & \multicolumn{3}{c}{Gaussian Privacy}                          \\\hline \hline
PFedMe & 0.49 $\pm$ 0.01 & 0.49 $\pm$ 0.02 & 0.49 $\pm$ 0.01 \\
PFedKM & 0.50 $\pm$ 0.02 & 0.50 $\pm$ 0.02 & 0.48 $\pm$ 0.02 \\
FedRep & 0.47 $\pm$ 0.03 & 0.50 $\pm$ 0.01 & 0.49 $\pm$ 0.01 \\
Global & 0.34 $\pm$ 0.01 & 0.34 $\pm$ 0.01 & 0.35 $\pm$ 0.03 \\
Global+ & 0.33 $\pm$ 0.00 & 0.33 $\pm$ 0.00 & 0.33 $\pm$ 0.00\\
FedEmbed-Personal & 0.49 $\pm$ 0.01 & 0.50 $\pm$ 0.02 & 0.49 $\pm$ 0.01 \\
FedEmbed-SOM & 0.50 $\pm$ 0.07 & 0.49 $\pm$ 0.05 & 0.45 $\pm$ 0.05 \\
\textbf{FedEmbed-Prototype} & 0.52 $\pm$ 0.21 & 0.55 $\pm$ 0.27 & \textbf{0.48 $\pm$ 0.13} \\
\textbf{FedEmbed-Type} & \textbf{0.54 $\pm$ 0.20} & \textbf{0.60 $\pm$ 0.14} & \textbf{0.48 $\pm$ 0.18}
\end{tabular}
\end{table*}

\newpage

\section{Speaker Preference Clustering Results}
\label{sec:appendix-clustering}

\begin{figure}[h]
    \centering
    \includegraphics[width=\linewidth]{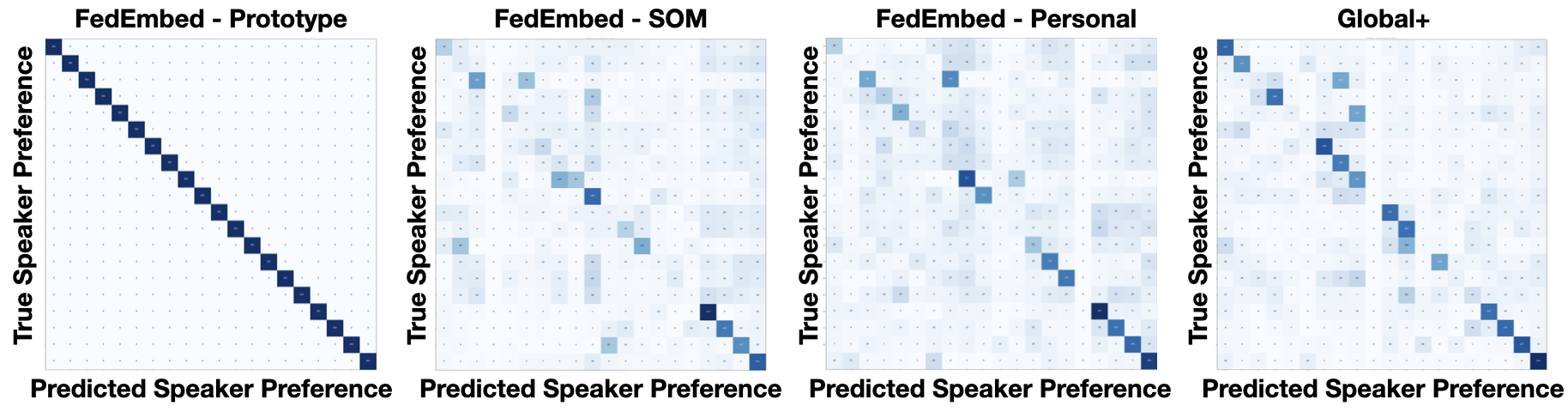}
    \caption{Confusion matrices for ground-truth user preferences and predicted user-preferences using personal embeddings derived from our balanced speaker preference dataset. \textit{FedEmbed-Prototype} perfectly recovers user preferences, but other approaches suffer from significant confusion on this more challenging dataset.}
    \label{fig:speaker-cluster-plot}
\end{figure}

To better understand personalization performance on our more challenging speaker-preference dataset, we present preference-assignment confusion matrices in Figure \ref{fig:speaker-cluster-plot}. From these confusion matrices, we observe that the FedEmbed-Prototype approach is able to accurately separate client preferences in this domain. Other approaches instead exhibit significant error, often recovering only a handful of preferences and showing significant confusion across the board. 


\end{document}